\documentclass{article}

\PassOptionsToPackage{numbers}{natbib}

\usepackage[preprint]{nips_2018}

\usepackage[utf8]{inputenc} 
\usepackage[T1]{fontenc}    
\usepackage{hyperref}       
\usepackage{url}            
\usepackage{booktabs}       
\usepackage{amsfonts}       
\usepackage{nicefrac}       
\usepackage{microtype}      
\usepackage{amsmath}
\usepackage{times}
\usepackage[psamsfonts]{amssymb}
\usepackage{amstext}
\usepackage{amsthm}
\usepackage{latexsym}
\usepackage{color}
\usepackage{graphicx}
\usepackage{enumerate}
\usepackage{algorithm}
\usepackage{url}
\usepackage{dsfont}
\usepackage[noend]{algpseudocode}

\usepackage{multirow}
\usepackage{makecell}

\usepackage{subcaption}
\usepackage{wrapfig}
\usepackage{changepage}
\usepackage{float}
\usepackage{bm}

\usepackage{pgfplots}
\pgfplotsset{compat=1.8}
\usepgfplotslibrary{statistics}

\usepackage{tikz} 
\usetikzlibrary{fit,calc,positioning,decorations.pathreplacing,matrix,arrows.meta}

\usepackage{relsize}

\title{Efficient Entropy for Policy Gradient with Multidimensional Action Space}

\author{
Yiming Zhang$^\dag$ \hspace{3mm} Quan Ho Vuong$^\ddag$  \hspace{3mm} Kenny Song$^{\S}$ \hspace{3mm} Xiao-Yue Gong$^{\P}$ \hspace{3mm} Keith W. Ross$^{\dag\S}$\\
$^\dag$ New York University\\
$^\ddag$ New York University Abu Dhabi \\
$\S$ New York University Shanghai\\
$\P$ Massachusetts Institute of Technology \\ 
\texttt{yiming.zhang@cs.nyu.edu, quan.vuong@nyu.edu, kenny.song@nyu.edu} \\
\texttt{xygong@mit.edu, keithwross@nyu.edu} \\
}

\newcommand{\E}{\mathrm{E}}

\newcommand{\Expect}{{\rm I\kern-.3em E}}
\DeclareMathOperator*{\argmax}{\rm argmax}

\newcommand{\prob}{p_{\theta}}
\newcommand{\entropy}{H_{\theta}}

\newcommand{\entropym}{\widetilde{H}_{\theta}}
\newcommand{\entropyx}{\widehat{H}_{\theta}}

\newcommand{\mat}[1]{{\mathbf #1}}

\newcommand{\A}{\mat{A}}

\newcommand{\act}{\mat{a}}
\newcommand{\cA}{\mathcal{A}}

\newcommand{\cS}{\mathcal{S}}

\newcommand{\bsigma}{{\boldsymbol \Sigma}}
\newcommand{\bmu}{{\boldsymbol \mu}}

\newcommand{\derp}{\nabla_\theta}
\newcommand{\lp}{\log \prob}
\newcommand{\sumtod}[1]{\sum_{#1=1}^d}
\newcommand{\EofA}{\E_{\A \sim \prob}}

\newcommand{\lpA}{\lp (A_i | \A_{i-1})}
\newcommand{\lpAj}{\lp (A_j | \A_{j-1})}

\newtheorem{theorem}{Theorem}

\begin{document}

\maketitle

\begin{abstract}
In recent years, deep reinforcement learning has been shown to be adept at solving sequential decision processes with high-dimensional state spaces such as in the Atari games. Many reinforcement learning problems, however, involve high-dimensional discrete action spaces as well as high-dimensional state spaces. This paper considers entropy bonus, which is used to encourage exploration in policy gradient. In the case of high-dimensional action spaces, calculating the entropy and its gradient requires enumerating all the actions in the action space and running forward and backpropagation for each action, which may be computationally infeasible. We develop several novel unbiased estimators for the entropy bonus and its gradient. We apply these estimators to several models for the parameterized policies, including Independent Sampling, CommNet, Autoregressive with Modified MDP, and Autoregressive with LSTM. Finally, we test our algorithms on two environments: a multi-hunter multi-rabbit grid game and a multi-agent multi-arm bandit problem. The results show that our entropy estimators substantially improve performance with marginal additional computational cost. 
\end{abstract}

\section{Introduction}

In recent years, deep reinforcement learning has been shown to be adept at solving sequential decision processes with high-dimensional state spaces such as in the Go game \cite{alpha_go} and Atari games \cite{original_atari_paper, human_level_control, multitask_learning_model_based_rl, combine_pg_dqn, actor_mimic_multi_task_transfer_learning_rl, sample_efficient_actor_critic, sobolev_training}.
In all of these success stories, the size of the action space was relatively small. 
Many Reinforcement Learning (RL) problems, however, involve high-dimensional action spaces as well as high-dimensional state spaces. Examples include StarCraft \cite{deepmind_starcraft, facebook_starcraft}, where there are many agents, each of which can take a finite number of actions; and coordinating self-driving cars at an intersection, where each car can take a finite set of actions \cite{comm_net}.

In policy gradient, in order to encourage sufficient exploration, an entropy bonus term is typically added to the objective function. However, in the case of high-dimensional action spaces, calculating the entropy and its gradient requires enumerating all the actions in the action space and running forward and backpropagation for each action, which may be computationally infeasible.

In this paper, we develop several novel unbiased estimators for the entropy bonus and its gradient. We apply these estimators to several models for the parameterized policies, including Independent Sampling, CommNet, Autoregressive with Modified MDP, and Autoregressive with LSTM. For all of these parameterizations, actions can be efficiently sampled from the policy distribution, and backpropagation can be employed for training. These parameterizations can be combined with the entropy bonus estimators and stochastic gradient descent, giving a new class of policy gradient algorithms with desirable exploration. Finally, we test our algorithms on two environments: a multi-hunter multi-rabbit grid game and a multi-agent multi-arm bandit problem. The results show that our entropy estimators can substantially improve performance with marginal additional computational cost. 

\section{Preliminaries}

Consider a Markov Decision Process (MDP) with a $d$-dimensional action space $\cA = \cA_{1} \times \cA_{2} \times \dots \times \cA_{d}$. 
Denote $\act = (a_1,\ldots,a_d)$ for an action in $\cA$. A policy $\pi(\cdot |s)$ specifies for each state $s$ a distribution over the action space $\cA$. In the standard RL setting, an agent interacts with an environment over a number of discrete time steps \cite{sutton_book, silver_lectures}. At time step $t$, the agent is in state $s_t$ and samples an action $\act_t$ from the policy distribution $\pi(\cdot|s_t)$. The agent then receives a scalar reward $r_t$ and the environment enters the next state $s_{t+1}$. The agent then samples $\act_{t+1}$ from $\pi(\cdot|s_{t+1})$ and so on. The process continues until the end of the episode, denoted by $T$.  The return $R_t = \sum_{k=0}^{T-t} \gamma^k r_{t+k}$ is the discounted accumulated return from time step $t$ until the end of the episode where $\gamma\in (0,1]$. 

In policy gradient, we consider a set of parameterized policies $\pi_\theta(\cdot|s)$, $\theta \in \Theta$, and attempt to find a good $\theta$ within a parameter set $\Theta$. Typically, the policy $\pi_\theta(\cdot|s)$ is generated by a neural network with $\theta$ denoting the network's weights and biases. The parameters $\theta$ are updated by performing stochastic gradient ascent on the expected reward. One example of such an algorithm is REINFORCE \cite{william}, where in a given episode at time step $t$, $\theta$ are updated as followed:
$$
\Delta \theta = \alpha \sum_{t=0}^T \nabla_\theta \log \pi_\theta(\act_t | s_t) (R_t - b_t(s_t))\label{eq:pg_update}
$$
where $b_t(s_t)$ is a baseline. It is well known that the policy gradient algorithm often converges to a local optimum. To discourage convergence to a highly suboptimal policy, the policy entropy is typically added to the update rule:
\begin{equation}
	\label{eqn:reinforce|ment_update_rule}
	\Delta \theta = \alpha \sum_{t=0}^T  [\nabla_\theta \log \pi_\theta(\act_t | s_t) (R_t - b_t(s_t)) + \beta \nabla_\theta H_\theta (s_t)]
\end{equation}
where
\begin{equation}
H_\theta (s_t) := - \sum_{\act \in \cA} \pi_\theta (\act| s_t) \log \pi_\theta (\act| s_t)
\label{eqn:PG_entropy}
\end{equation}
This approach is often referred to as adding entropy bonus or entropy regularization \cite{william} and is widely used in different applications, such as optimal control in Atari games \cite{async_rl}, multi-agent games \cite{multi_agent_openai} and  optimizer search for supervised machine learning with RL \cite{optimizer_search}. $\beta$ is referred to as the entropy weight.

\section{Policy Parameterization for Multidimensional Action Space}
For problems with discrete action spaces, policies are commonly parameterized as a feed-forward neural network (FFN) with a softmax output layer of dimension $|\cA|$. Therefore sampling from such a policy requires O($|\cA|$) effort. For multidimensional action spaces, $|\cA|$ grows exponential with the number of dimensions .

In order to efficiently sample from our policy, we consider an autoregressive model which can be sampled from each dimension sequentially. In our discussion, we will assume $|\cA_1|=\cdots=|\cA_d|=K$. To handle action sets of different sizes, we will include inconsequential actions. Here we review two such models, and note that sampling from both models only require summing over $O(dK)$ effort as opposed to $O(K^d)$ effort. We emphasize that our use of an autoregressive model to create multi-dimensional probability distributions is not novel. However, we need to provide a brief review to motivate our entropy calculation algorithms. 

\subsection{Using an LSTM to Generate the Parameterized Policy}

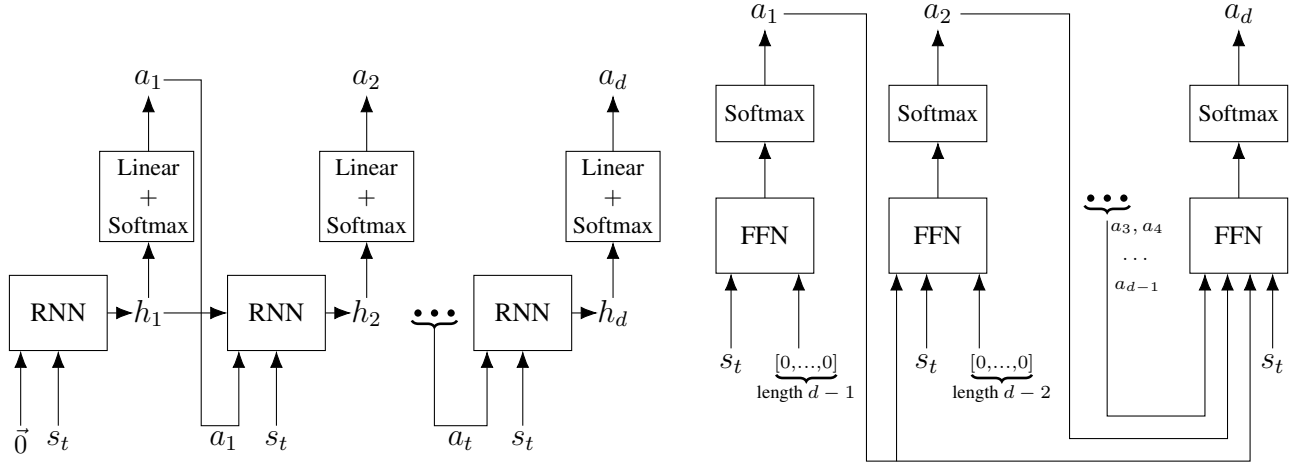
\begin{figure*}
\makebox[\textwidth]{
\subcaptionbox{The RNN architecture. To generate $a_i$, we input $s_t$ and $a_{i-1}$ into the RNN and then pass the resulting hidden state $h_i$ through a linear layer and a softmax to generate a distribution, from which we sample $a_i$.}
{\begin{subfigure}{.4\paperwidth}
	\begin{tikzpicture}[decoration={brace}]
	\draw [black] (1.75,2) rectangle (3.05,3);
	\node at (2.4, 2.5) {RNN};
	
	\draw [-{Latex[length=2.3mm]}] (1.9, 1) -- (1.9, 2);
	\node at (1.9, 0.8) {$\vec{0}$};

	\draw [-{Latex[length=2.3mm]}] (2.4,1) -- (2.4,2);
	\node at (2.4, 0.8) {$\mathlarger{\mathlarger{s_t}}$};
	
	\draw [-{Latex[length=2.3mm]}] (3.6, 2.7) -- (3.6,  3.45);
	\node at (3.6, 4.05) [align=center] {\footnotesize Linear \\ \footnotesize $+$ \\ \footnotesize Softmax};
	\draw [black] (2.95, 3.45) rectangle (4.2, 4.65);
	
	\draw [-{Latex[length=2.3mm]}] (3.6, 4.65) -- (3.6, 5.4);
	\node at (3.6, 5.6) {$\mathlarger{\mathlarger{a_1}}$};	
	\draw [-{Latex[length=2.3mm]}] (3.8, 5.6) -- (4.3, 5.6) -- (4.3, 1) -- (4.8, 1) -- (4.8, 2);
	\node at (4.6, 0.8) {$\mathlarger{\mathlarger{a_1}}$};

	\draw [-{Latex[length=2.3mm]}] (3.05,2.5) -- (3.4,2.5);
	\node at (3.6, 2.5) {$\mathlarger{\mathlarger{h_1}}$};
	\draw [-{Latex[length=2.3mm]}] (3.8,2.5) -- (4.65,2.5);
		
	\draw [black] (4.65,2) rectangle (5.95,3);
	\node at (5.3, 2.5) {RNN};
	
	\draw [-{Latex[length=2.3mm]}] (5.3,1) -- (5.3,2);
	\node at (5.3, 0.8) {$\mathlarger{\mathlarger{s_t}}$};
	
	\draw [-{Latex[length=2.3mm]}] (6.5, 2.7) -- (6.5,  3.45);
	\node at (6.5, 4.05) [align=center] {\footnotesize Linear \\ \footnotesize $+$ \\ \footnotesize Softmax};
	\draw [black] (5.875, 3.45) rectangle (7.125, 4.65);
	
	\draw [-{Latex[length=2.3mm]}] (6.5, 4.65) -- (6.5, 5.4);
	\node at (6.5, 5.6) {$\mathlarger{\mathlarger{a_2}}$};	
	
	\draw [-{Latex[length=2.3mm]}] (5.95, 2.5) -- (6.3, 2.5);
	\node at (6.5, 2.5) {$\mathlarger{\mathlarger{h_2}}$};

	\node at (7.4, 2.5) {\Huge ...};	
%
	\draw [black] (7.925, 2) rectangle (9.225, 3);
	\node at (8.575, 2.5) {RNN};
	
	\draw [decorate, line width=1pt] (7.7, 2.4) -- (7.1, 2.4);
	\draw [-{Latex[length=2.3mm]}] (7.4, 2.3) -- (7.4, 1) -- (8.1, 1) -- (8.1, 2);
	\node at (7.75, 0.8) {$\mathlarger{\mathlarger{a_t}}$};
	
	\draw [-{Latex[length=2.3mm]}] (8.575, 1) -- (8.575, 2);
	\node at (8.575, 0.8) {$\mathlarger{\mathlarger{s_t}}$};
	
	\draw [-{Latex[length=2.3mm]}] (9.225, 2.5) -- (9.575, 2.5);
	\node at (9.775, 2.5) {$\mathlarger{\mathlarger{h_d}}$};

	\draw [-{Latex[length=2.3mm]}] (9.775, 2.7) -- (9.775,  3.45);
	\node at (9.775, 4.05) [align=center] {\footnotesize Linear \\ \footnotesize $+$ \\ \footnotesize Softmax};
	\draw [black] (9.15, 3.45) rectangle (10.4, 4.65);
	
	\draw [-{Latex[length=2.3mm]}] (9.775, 4.65) -- (9.775, 5.4);
	\node at (9.775, 5.6) {$\mathlarger{\mathlarger{a_d}}$};	

\end{tikzpicture}
\end{subfigure}}
\hspace{0.04\textwidth}
\subcaptionbox{The MMDP architecture. To generate $a_i$, we input $s_t$ and $a_1, a_2, \dots , a_{i-1}$ into a FFN. The output is passed through a softmax 
layer, providing a distribution from which we sample $a_i$. Since the input size of the FFN is fixed, when generating $a_i$, constants $0$ serve as placeholders for $a_{i+1}, \dots, a_{d-1}$ in the input to the FFN.}
{\begin{subfigure}{.35\paperwidth}
	\begin{tikzpicture}[decoration={brace}]
	\draw [black] (2,2) rectangle (3.3,3);
	\node at (2.65, 2.5) {FFN};
	
	\draw [-{Latex[length=2.3mm]}] (2.2,1) -- (2.2,2);
	\node at (2.2, 0.8) {$\mathlarger{\mathlarger{s_t}}$};
	
	\draw [-{Latex[length=2.3mm]}] (3.1,1) -- (3.1,2);
	\node at (3.2, 0.8) [align=center] {$\scriptstyle [0, \dots, 0]$};
	\draw [decorate, line width=1pt] (3.6,0.65) -- (2.8, 0.65);
	\node at (3.2, 0.4) {\scriptsize length $d-1$};

	\draw [-{Latex[length=2.3mm]}] (2.65,3) -- (2.65,3.75);

	\node at (2.65, 4.125) {\footnotesize Softmax};
	\draw [black] (2, 3.75) rectangle (3.3,4.5);
	
	\draw [-{Latex[length=2.3mm]}] (2.65,4.5) -- (2.65,5.25);

	\node at (2.65, 5.45) {$\mathlarger{\mathlarger{a_1}}$};	
	
	\draw [-{Latex[length=2.3mm]}] (2.9, 5.45) -- (4, 5.45) -- (4, -0.5) -- (4.4, -0.5) -- (4.4, 2);
	
	\draw [-{Latex[length=2.3mm]}] (4.4, -0.5) -- (9.1, -0.5) -- (9.1, 2);

	\draw [black] (4.3,2) rectangle (5.6,3);
	\node at (4.95, 2.5) {FFN};
	
	\draw [-{Latex[length=2.3mm]}] (4.8,1) -- (4.8,2);
	\node at (4.8, 0.8) {$\mathlarger{\mathlarger{s_t}}$};
	
	\draw [-{Latex[length=2.3mm]}] (5.5,1) -- (5.5,2);
	\node at (5.8, 0.8) [align=center] {$\scriptstyle [0, \dots, 0]$};
	\draw [decorate, line width=1pt] (6.2, 0.65) -- (5.4, 0.65);
	\node at (5.8, 0.4) {\scriptsize length $d-2$};

	\draw [-{Latex[length=2.3mm]}] (4.95,3) -- (4.95,3.75);

	\node at (4.95, 4.125) {\footnotesize Softmax};
	\draw [black] (4.3, 3.75) rectangle (5.6,4.5);
	
	\draw [-{Latex[length=2.3mm]}] (4.95,4.5) -- (4.95,5.25);

	\node at (4.95, 5.45) {$\mathlarger{\mathlarger{a_2}}$};
	
	\draw [-{Latex[length=2.3mm]}] (5.25, 5.45) -- (6.7, 5.45) -- (6.7, -0.2) -- (8.8, -0.2) -- (8.8, 2);

	\node at (7.2, 3) {\Huge ...};
	\draw [decorate, line width=1pt] (7.5, 2.9) -- (6.9, 2.9);
	\node at (7.6, 2.2) [align=center] {\scriptsize $a_3, a_4$ \\ \scriptsize $\dots$ \\ \scriptsize $a_{d-1}$};
	
	\draw [-{Latex[length=2.3mm]}] (7.2, 2.7) -- (7.2, 0.1) -- (8.5, 0.1) -- (8.5, 2);

	\draw [black] (8.3,2) rectangle (9.6,3);
	\node at (8.95, 2.5) {FFN};
	
	\draw [-{Latex[length=2.3mm]}] (9.4,1) -- (9.4,2);
	\node at (9.4, 0.8) {$\mathlarger{\mathlarger{s_t}}$};

	\draw [-{Latex[length=2.3mm]}] (8.95,3) -- (8.95,3.75);
		
	\node at (8.95, 4.125) {\footnotesize Softmax};
	\draw [black] (8.3, 3.75) rectangle (9.6, 4.5);
	
	\draw [-{Latex[length=2.3mm]}] (8.95,4.5) -- (8.95,5.25);

	\node at (8.95, 5.45) {$\mathlarger{\mathlarger{a_d}}$};
	
\end{tikzpicture}
\end{subfigure}}

}
\caption{The RNN and MMDP architectures for generating parameterized policies.}

\label{fig:lstm_mmdp}
\end{figure*}

LSTMs have recently been used with great success for autoregressive models in language translation tasks \cite{lstm_translation}. An LSTM can also be used to create a parameterized multi-dimensional distribution $p_\theta(\cdot)$ and to sample $\act$ from that distribution (\autoref{fig:lstm_mmdp}(a)). To generate $a_i$, we run a forward pass through the LSTM with the input being $a_{i-1}$ and the current state $s_t$ (and implicitly on $a_1,\ldots,a_{i-1}$ which influences $h_{i-1}$). This produces a hidden state $h_i$, which is then passed through a linear layer, producing a $K$ dimensional vector. The softmax of this vector is taken to produce the one-dimensional conditional distribution $p_\theta (a| \act_{i-1})$, $a \in \cA_i$. $a_i$ is sampled from this one-dimensional distribution, and is then fed into the next stage of the LSTM to produce $a_{i+1}$. We note that this approach is an adaptation of sequence modeling in supervised machine learning \cite{wave_net} to reinforcement learning and has also been proposed by \cite{google_sdqn_cont_action} and \cite{actor_critic_sequence_prediction}.

\subsection{Using MMDP to Generate Parameterized Policy}

The underlying MDP can be modified to create an equivalent MDP for which the action space is one-dimensional. We refer to this MDP as the Modified MDP (MMDP). 
In the original MDP, we have state space $\cS$ and action space $\cA = \cA_{1} \times \cA_{2} \times \dots \times \cA_{d}$. In MMDP, the state encapsulates the original state and all the action dimensions selected for state $s$ so far (\autoref{fig:lstm_mmdp}(b)).
We note that \cite{google_sdqn_cont_action} recently and independently proposed the reformulation of the MDP into MMDP.

\section{Entropy Bonus Approximation for Multidimensional Action Space}

As shown in (\ref{eqn:reinforce|ment_update_rule}), an entropy bonus is typically included to enhance exploration. However, for large multi-dimensional action space, calculating the entropy and its gradient requires enumerating all the actions in the action space and running forward and backpropagation for each action. In this section, we develop computationally efficient unbiased estimates for the entropy and its gradient. These computationally efficient algorithms can be combined with the autoregressive models discussed in the previous section to provide end-to-end computationally efficient schemes. 

To abbreviate notations, we write $p_\theta(\act)$ for $\pi_\theta(\act | s_t)$ and $\act_i$ for $(a_1, a_2,\ldots, a_i)$. We consider auto-regressive models whereby the sample components $a_i$, $i=1,\ldots, d$ are sequentially generated. In particular, after obtaining $a_1, a_2,\ldots, a_{i-1}$, we will generate $a_i \in \cA_i$ from some parameterized distribution $p_\theta (\cdot| \act_{i-1})$ defined over the one-dimensional set $\cA_i$.  After generating the distribution
$p_\theta (\cdot| \act_{i-1})$, $i=1,\ldots, d$ and the action components $a_1,\ldots,a_d$ sequentially, we then define 
$p_\theta(\act) = \prod_{i=1}^d p_\theta (a_i | \act_{i-1})$. 

Let $\A = (A_1,\dots,A_d)$ denote a random variable with distribution $\prob(\cdot)$.  Let $\entropy$ denote the exact entropy of the distribution $\prob(\act)$:
\begin{align*}
	\entropy & = -\sum_{\act}\prob(\act)\log\prob(\act) \\
	& = - \EofA [\log\prob(\A)] \\
	& = -\sum_{i=1}^d \EofA [\log\prob(A_i|\A_{i-1})] \label{entropy_expanded}
\end{align*}

\subsection{Crude Unbiased Estimator}
\label{crude_approximation}
During training within an episode, for each state $s_t$, the policy generates an action $\act$. We refer to this generated action as the episodic sample. A crude approximation of the entropy bonus is:
\begin{displaymath}
H_\theta^{\text{crude}}(\act) = - \log \prob(\act)
= - \sum_{i=1}^{d} \log \prob (a_i | \act_{i-1})
\end{displaymath}
This approximation is an unbiased estimate of $H_\theta$ but its variance is likely to be large. To reduce the variance, we can generate multiple action samples when in $s_t$ and average the log action probabilities over the samples. However, generating a large number of samples is costly, especially when each sample is generated from a neural network, since each sample requires one additional forward pass. 

\subsection{Smoothed Estimator}
\label{approximate_entropy}

This section proposes an alternative unbiased estimator for $H_\theta$ which only requires the one episodic sample and accounts for the entropy along each dimension of the action space:
\begin{align*}
\entropym(\act) & := - \sum_{i=1}^d \sum_{{a \in \cA_i}} \prob(a | \act_{i-1}) \log p_\theta(a| \act_{i-1}) \\
           & = \sum_{i=1}^d H_\theta^{(i)} (\act_{i-1})
\end{align*}
where
\begin{displaymath}
H_\theta^{(i)} (\act_{i-1}) := - \sum_{{a \in \cA_i}} \prob(a | \act_{i-1}) \log p_\theta(a| \act_{i-1})
\end{displaymath}
which is the entropy of $A_i$ conditioned on $\act_{i-1}$. This estimate of the entropy bonus is computationally efficient since for each dimension $i$, we would need to obtain $p_\theta(\cdot | \act_{i-1})$, its log and gradient anyway during training. We refer to this approximation as the smoothed entropy.

The smoothed estimate of the entropy $\entropym(\A)$ has several appealing properties. The proofs of Theorem 1 and Theorem 3 are straightforward and omitted. 
\begin{theorem}
$\entropym(\A)$ is an unbiased estimator of the exact entropy $\entropy$.
\end{theorem}

\begin{theorem}
	If $\prob(\act)$ has a multivariate normal distribution with mean and variance depending on $\theta$, then: 
	$$\entropym(\act) = \entropy \quad \forall \act \in \cA$$
	\\
Thus, the smoothed estimate of the entropy equals the exact entropy for a multivariate normal parameterization of the policy.    
\label{a:smooth_equal_exact}
\end{theorem}
See Appendix B for proof.

\begin{theorem}
(i) If there exists a sequence of weights $\theta_1,\theta_2,\dots$ in $\Theta$ such that $p_{\theta_n}(\cdot)$ converges to the uniform distribution over $\cA$, then
\[
\sup_{\theta \in \Theta}\entropym(\act) = \sup_{\theta \in \Theta} \entropy \quad \forall \act \in \cA
\] \\
(ii) If there exists a sequence of weights $\theta_1,\theta_2,\dots$ in $\Theta$ such that $p_{\theta_n}(\act^*)\rightarrow 1$ for some $\act^*$, then
\[
\inf_{\theta \in \Theta}\entropym(\act) = \inf_{\theta \in \Theta}\entropy = 0 \quad \forall \act \in \cA
\]
Thus, the smoothed estimate of the entropy $\entropym(\act)$ mimics the exact entropy in that it has the same supremum and infinum values as the exact entropy.
\end{theorem}

The above theorems indicate that $\entropym(\act)$ may serve as a good proxy for $\entropy$: it is an unbiased estimator for $\entropy$, it has the same minimum and maximum values when varying $\theta$; and in the special case when $\prob(\act)$ has a multivariate normal distribution, it is actually equal to $\entropy$ for all $\act\in\cA$. Our numerical experiments have shown that the smoothed estimator $\entropym(\act)$ typically has lower variance than the crude estimator $H_\theta^{\text{crude}}(\act)$. However, it is not generally true that the smoothed estimator always has lower variance as counterexamples can be found.

\subsection{Smoothed Mode Estimator}
\label{a:smooth_mode_entropy}

For the smoothed estimate of the entropy $\entropym(\act)$, we use the episodic action $\act$ to form the estimate. We now consider alternative choices of actions which may improve performance at modest additional computational cost. First consider $E^*=-\sum_{i=1}^d\sum_{a\in\cA_i}\prob(a|a^*_1,\dots,a^*_{i-1})\log\prob(a|a^*_1,\dots,a^*_{i-1})$ where $\act^*=(a^*_1,\dots,a^*_d)=\argmax_{\act\in\cA}\prob(\act)$. Thus in this case, instead of calculating the smoothed estimate of the entropy with the episodic action $\act$, we calculate it with the most likely action $\act^*$. The problem here is that it is not easy to find $\act^*$ when
the given conditional probabilities $\prob(a_i|\act_{i-1})$ are not in closed form but only available algorithmically as outputs of neural networks. A more computationally efficient approach would be to choose the action greedily: 
$\hat{a}_1 = \argmax_{a\in\cA_1}\prob(a)$ and $\hat{a}_i = \argmax_{a\in\cA_i}\prob(a|\hat{a}_1,\dots,\hat{a}_{i-1})$ for $i=2,\ldots,d$. This leads to the definition $\entropyx = -\sum_{i=1}^d\sum_{a\in\cA_i}\prob(a|\hat{a}_1,\dots,\hat{a}_{i-1})\log\prob(a|\hat{a}_1,\dots,\hat{a}_{i-1})$. The action $\hat{\act}$ is an approximation for the mode of the distribution $\prob(\cdot)$. As often done in NLP, we can use beam search to determine an action $\act'$ that has higher probability, that is, $\prob(\act') \geq \prob(\hat{\act})$.  Indeed,  the above $\entropyx$ definition is beam search with beam size equal to 1. We refer to $\entropyx$ as the smoothed mode estimate.

$\entropyx$ with an appropriate beam size may be a better approximation for the entropy $\entropy$ than $\entropym(\act)$. However, calculating $\entropyx$ and its gradient comes with some computational cost. For example, with a beam size equal to one, we would have to make two passes through the policy neural network at each time step: one to obtain the episodic sample $\act$ and the other to obtain the greedy action $\hat{\act}$. For beam size $n$, we would need to make $n+1$ passes. We note that $\entropyx$  is a biased estimator for $\entropy$ but with no variance. Thus there is a bias-variance tradeoff between $\entropym(\act)$ and $\entropyx$. Note that $\entropyx$ also satisfies Theorems 2 and 3 in \autoref{approximate_entropy}.

\subsection{Estimating the Gradient of the Entropy}

So far we have been looking at estimates of entropy. But the update rule (\ref{eqn:reinforce|ment_update_rule}) uses the gradient of the entropy rather than the entropy. As it turns out, the gradients of the estimators $H_\theta^{\text{crude}}(\act)$ and $\entropym(\act)$ are not unbiased estimates of the gradient of the entropy. In this subsection, we provide unbiased estimators for the gradient of the entropy. For simplicity, in this section, we assume a one-step decision setting, such as in a multi-armed bandit problem. A straightforward calculation shows:
\begin{equation}
		\derp \entropy = \EofA [ - \lp (\A) \derp \lp (\A)]
\label{der_entropy_expansion}
\end{equation}
Suppose $\act$ is one sample from $\prob(\cdot)$. A crude unbiased estimator for the gradient of the entropy therefore is:
$- \lp (\act) \derp \lp (\act) = \lp (\act) \derp H_\theta^{\text{crude}}(\act)$. Note that this estimator is equal to the gradient of the crude estimator multiplied by a correction factor.  

Analogous to the smoothed estimator for entropy, we can also derive a smoothed estimator for the gradient of the entropy.
\begin{theorem}
If $\act$ is a sample from $\prob(\cdot)$, then
$$ \derp \entropym(\act) + \sum_{i=1}^d H_\theta^{(i)} (\act_{i-1}) 
\derp \sum_{j=1}^{i-1} \lp (a_j| \act_{j-1})$$
is an unbiased estimator for the gradient of the entropy.
\label{a:unbiased_entropy_gradient_theorem}
\end{theorem}
See Appendix C for proof. 

Note that this estimate for the gradient of the entropy is equal to the gradient of the smoothed estimate $\entropym(\act)$ plus a correction term. We refer to this estimate of the entropy gradient as the unbiased gradient estimate.

\section{Experimental Results}
\label{experiment_results}

We designed experiments to compare the different entropy estimators the LSTM, MMDP, and CommNet model, a related approach introduced by \cite{comm_net}. As a baseline, we will use the Independent Sampling (IS) model which is an FFN that takes as input the state, creates a representation of the state, and from the representation outputs $d$ softmax heads, from which the value of each action dimension can be sampled independently \cite{comm_net}. In this case, the smoothed estimate is equal to the exact entropy. For each entropy approximation, the entropy weight $\beta$ was tuned to give the highest reward. For IS and MMDP, the number of hidden layers was tuned from 1 to 7. For CommNet, the number of communication steps was tuned from 2 to 5, the learning rate was tuned between 3e-3 and 3e-4 and the size of the policy hidden layer was tuned between 128 and 256.  

\subsection{Hunters and Rabbits}
\label{hunter_rabbit}

In this environment, there is a $n \times n$ grid. At the beginning of each episode, $d$ hunters and $d$ rabbits are randomly placed in the grid. The rabbits remain fixed in the episode, and each hunter can move to a neighboring square (including diagonal neighbors) or stay at the current square. So each hunter has nine possible actions, and altogether there are $|\cA| = 9^d$ actions at each time step. When a hunter enters a square with a rabbit, the hunter captures the rabbit and remains there until the end of the episode. In each episode, the goal is for the hunters to capture the rabbits as quickly as possible. Each episode is allowed to run for at most ten thousands time steps.  

To provide a dense reward signal, we modify the goal as following: capturing a rabbit gives a reward of $1$, which is discounted by the number of time steps taken since the beginning of the episode. The discount factor is 0.8. The goal is to maximize the episode's total discounted reward. After a hunter captures a rabbit, they both become inactive. 

\textit{Comparison of different entropy estimates for IS, LSTM, MMDP and CommNet}

\autoref{tab:lstm_mmdp_performance} shows the performance of the IS, LSTM, MMDP and CommNet models with the different entropy estimates. Training and evaluation were performed in a square grid of 5 by 5 with 5 hunters and 5 rabbits. Results are averaged over 5 seeds. For each seed, training and evaluation were run for 1 million and 1 thousand episodes respectively.

As compared with no entropy, crude entropy can actually reduce performance. However, smoothed entropy and smoothed mode entropy always increase performance, often significantly. For the LSTM model, the best performing approximation is smoothed entropy, reducing the mean episode length by $45\%$ and increasing the mean episode reward by $10\%$ compared to without entropy. We also note that there is not a significant difference in performance between the smoothed entropy, smoothed mode entropy, and the unbiased gradient approaches. When comparing the four models, we see that the LSTM model with smoothed entropy does significantly better the other three models. The CommNet model could potentially be improved by allowing the hunters to see more of the state; this could be investigated in future research. 

\begin{table*}
    \caption{Performance of IS, LSTM, MMDP and CommNet across different entropy approximations.}
\centering
    \begin{tabular}{ | c | c | c | c | c | c |}
    \hline
    & \thead{Without \\ Entropy} & \thead{Crude \\ Entropy}  &  \thead{Smoothed \\ Entropy}  & \thead{Smoothed \\ Mode Entropy} & \thead{Unbiased Gradient \\ Estimate} \\ \hline \hline

	\thead{IS Mean \\ Episode Length} & 98.7 \small $\pm$ 78.9 & 32 \small $\pm$ 12.3 &   11.8 \small $\pm$ 1.9 &  11.8 \small $\pm$ 1.9  & 11.8 \small $\pm$ 1.9  \\ \hline
	
    \thead{LSTM Mean \\ Episode Length} & 10.1 \small $\pm$ 1.9 & 19 \small $\pm$ 8.7 & $\mathbf{5.6 \small \pm 0.2}$ & 6.0 \small $\pm$ 0.2 & 6.0 \small $\pm$ 0.1 \\ \hline
    
    \thead{MMDP Mean \\ Episode Length}  & 21.5 \small $\pm$ 3.7 & 37.3 \small $\pm $ 29.6 & 10.6 \small $\pm$ 0.7 & 10.6 \small $\pm$ 0.7 & 9.8 \small $\pm$ 0.6 \\\hline
    
    \thead{CommNet Mean \\ Episode Length}  & 22.7 \small $\pm$ 0.6 & 22.3 \small $\pm$ 0.4 & 21.9 \small $\pm$ 0.4 &  21.9 \small $\pm$ 0.4  &  21.9 \small $\pm$ 0.4 \\\hline \hline

    \thead{IS Mean \\ Episode Reward} & 2.2 \small $\pm$ 0.03 &  2.4 \small $\pm$ 0.05 &  2.7 \small $\pm$ 0.01 &  2.7 \small $\pm$ 0.01  & 2.7 \small $\pm$ 0.01 \\ \hline

    \thead{LSTM Mean \\ Episode Reward} & 3.0 \small $\pm$ 0.06 & 3.0 \small $\pm$ 0.03 & $\mathbf{3.3 \small \pm 0.04}$ &  3.2 \small $\pm$ 0.04 & 3.2 \small $\pm$ 0.02 \\ \hline
    
    \thead{MMDP Mean \\ Episode Reward} & 2.8 \small $\pm$ 0.03 & 2.7 \small $\pm$ 0.03 & 2.9 \small $\pm$ 0.03 & 2.8 \small $\pm$ 0.04 & 2.9 \small $\pm$ 0.02  \\ \hline
    
     \thead{CommNet Mean \\ Episode Reward} &  2.5 \small $\pm$ 0.01  &   2.6 \small $\pm$ 0.01 & 2.6 \small $\pm$ 0.01 & 2.6 \small $\pm$ 0.01 & 2.6 \small $\pm$ 0.01 \\ \hline

    \end{tabular}
    \label{tab:lstm_mmdp_performance}
\end{table*}

\begin{figure*}
\caption{IS, LSTM and MMDP results across 5 seeds (y-axis denotes mean episode length).}

\makebox[0.95\textwidth]{

\subcaptionbox{IS}
{\begin{subfigure}{0.18\paperwidth}
\begin{tikzpicture}

\begin{axis}[
    boxplot/draw direction=y,
    x axis line style={opacity=1},
    axis x line*=bottom,
    axis y line=left,
    enlarge y limits,
    ymajorgrids,
    xtick={1,2},
    xticklabels={Without, Smoothed},
    height=3.95cm,
]
\addplot [
        boxplot prepared={
            lower whisker=24.32, lower quartile=37.806,
            median=70.042,
            upper quartile=74.749, upper whisker=286.668,
        },
    ] coordinates {};
    \addplot [
        boxplot prepared={
            lower whisker=10.043, lower quartile=10.661,
            median=10.6,
            upper quartile=10.843, upper whisker=16.911,
        },
    ] coordinates {};
\end{axis}

\end{tikzpicture}
\end{subfigure}}

\subcaptionbox{LSTM}
{\begin{subfigure}{0.18\paperwidth}
\begin{tikzpicture}

\begin{axis}[
    boxplot/draw direction=y,
    x axis line style={opacity=1},
    axis x line*=bottom,
    axis y line=left,
    enlarge y limits,
    ymajorgrids,
    xtick={1,2},
    xticklabels={Without, Smoothed},
    height=3.85cm,
]
\addplot [
        boxplot prepared={
            lower whisker=7.57, lower quartile=8.857,
            median=9.285,
            upper quartile=10.713, upper whisker=14.195,
        },
    ] coordinates {};
    \addplot [
        boxplot prepared={
            lower whisker=5.351, lower quartile=5.419,
            median=5.459,
            upper quartile=5.585, upper whisker=5.996,
        },
    ] coordinates {};
\end{axis}

\end{tikzpicture}
\end{subfigure}}

\subcaptionbox{MMDP}
{\begin{subfigure}{0.18\paperwidth}
\begin{tikzpicture}

\begin{axis}[
    boxplot/draw direction=y,
    x axis line style={opacity=1},
    axis x line*=bottom,
    axis y line=left,
    enlarge y limits,
    ymajorgrids,
    xtick={1,2},
    xticklabels={Without, Smoothed},
    height=4.05cm,
]
\addplot [
        boxplot prepared={
            lower whisker=15.164, lower quartile=18.876,
            median=20.264,
            upper quartile=26.638, upper whisker=26.717,
        },
    ] coordinates {};
    \addplot [
        boxplot prepared={
            lower whisker=9.691, lower quartile=9.783,
            median=10.643,
            upper quartile=11.118, upper whisker=11.773,
        },
    ] coordinates {};
\end{axis}

\end{tikzpicture}
\end{subfigure}}

}

\label{fig:lstm_robust_over_seeds}
\end{figure*}
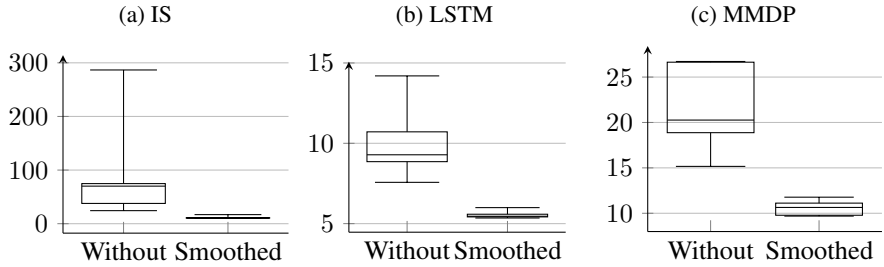

The smoothed estimator is also more robust with respect to the initial seed than without entropy as shown in \autoref{fig:lstm_robust_over_seeds}. For example, for the LSTM model, in the case of without entropy, seed 0 leads to significantly worse results than the seeds 1-4. This does not happen with the smoothed estimator. 

\textit{Entropy approximations versus exact entropy}

We now consider how policies trained with entropy approximations compare with polices trained with exact entropy. In order to calculate exact entropy in an acceptable amount of time, we reduced the number of hunters and rabbits to 4 hunters and 4 rabbits. Training was run for 50,000 episodes.
\autoref{tab:vs_exact_entropy} shows the performance differences between policies trained with entropy approximations and exact entropy.  We see that the best entropy approximations perform only slightly worse than exact entropy for both LSTM and MMDP. Once again we see that the LSTM model performs better than the MMDP model.

\begin{table*}
    \caption{LSTM and MMDP results for entropy approximation versus exact entropy.}
\begin{adjustwidth}{-0.15cm}{}
\centering
    \begin{tabular}{ | c  | c | c | c | c |}
    \hline
     &  \thead{LSTM \\ Smoothed Entropy} & \thead{LSTM \\ Exact Entropy} & \thead{MMDP Unbiased \\ Gradient Estimate} & \thead{MMDP \\ Exact Entropy}\\ \hline
    \thead{Mean Episode Length} & 9.0 \small $\pm$ 0.3 & $\mathbf{8.9 \small \pm 0.2}$ & 11.5 \small $\pm$ 0.3 & 10.7 \small $\pm$ 0.4 \\ \hline
    \thead{Mean Episode Reward} & 2.14 \small $\pm$ 0.02 & $\mathbf{2.19 \small \pm 0.02}$ & 2.01 \small $\pm$ 0.01 & 2.1 \small $\pm$ 0.01 \\ \hline
    \end{tabular}
\end{adjustwidth}    
    \label{tab:vs_exact_entropy}
\end{table*}

\subsection{Multi-agent Multi-arm Bandits}

We examine a multi-agent version of the standard multi-armed bandit problem, where there are $d$ agents each pulling one of $K$ arms, with $d\leq K$. The $k^{th}$ arm generates a reward $r_k$. The total reward in a round is generated as follows. In each round, each agent chooses an arm. All of the chosen arms are then pulled, with each pulled arm generating a reward. Note that the total number of arms chosen, $c$, may be less than $d$ since some agents may choose the same arm. The total reward is the sum of rewards from the $c$ chosen arms. The optimal policy is for the $d$ agents to collectively pull the $d$ arms with the highest rewards. Additionally, among all the optimal assignments of $d$ agents to the $d$ arms that yield the highest reward, we add a bonus reward with probability $p^*$ if one particular agents-to-arms configuration is chosen.

We performed experiments with 4 agents and 10 arms, with the $k^{th}$ arm providing a reward of $k$. The exceptional assignment gets a bonus of 166 (making a total reward of 200) with probability 0.01, and no bonus with probability 0.99. Thus the maximum expected reward is 35.66. Training was run for 100,000 rounds for each of 10 seeds. \autoref{tab:lstm_performance_bandits} shows average results for the last 500 of the 100,000 rounds. 

\begin{table*}
\centering
    \caption{Performance of IS and LSTM policy parameterizations.}
    \begin{tabular}{ | c  | c | c | c | c |}
    \hline
    & \thead{Without \\ Entropy} & \thead{Crude \\ Entropy}  & \thead{Smoothed \\ Entropy} & \thead{Unbiased\\ Gradient Estimate} \\ \hline
    \thead{IS Mean Reward} & 34.2 \small $\pm$ 1.3 & 34.4 \small $\pm$ 1.3 &  34.2 \small $\pm$ 1.3 & $34.2 \pm 1.3$ \\ \hline
        \thead{LSTM Mean Reward} & 34.9 \small $\pm$ 0.8 & 35.5 \small $\pm$ 1.1 &  35.9 \small $\pm$ 0.8 & $\mathbf{35.9 \small \pm 0.3}$ \\ \hline
        \thead{IS Percentage \\ Optimal Config Pulled} & 19.8 \small $\pm$ 39.7 & 29.7 \small $\pm$ 49.6 & 19.7 \small $\pm$ 39.7 & $19.7 \small \pm 39.7$ \\ \hline
      \thead{LSTM Percentage \\ Optimal Config Pulled} & 39.8 \small $\pm$ 35.9 & 59.4 \small $\pm$ 35.7 & 95.0 \small $\pm$ 1.9 & $\mathbf{95.7 \small \pm 2.7}$ \\ \hline
    \end{tabular}
    \label{tab:lstm_performance_bandits}
\end{table*}

The results for the multi-agent bandit problem are consistent with those for the hunter-rabbit problem. Policies obtained with the entropy approximations all perform better than policies obtained without entropy or with crude entropy, particularly for the percentage of rounds the arms are pulled with the optimal configuration. Note that LSTM with the unbiased gradient estimator gives the best results.

\section{Related Work}
\label{related_work}

Metz et al. \cite{google_sdqn_cont_action} recently and independently proposed the reformulation of MDP into the MMDP and the LSTM policy parameterization. They inject noise into the action space to encourage exploration. Usunier et al. \cite{sc_episodic_explore} uses MMDP and noise injection in the parameter space to achieve high performance in multi-agent Starcraft micro-management tasks. Instead of noise injection, we propose novel estimators for the entropy bonus that is often used to encourage exploration in policy gradient. 

While entropy regularization has been mostly used in policy gradient algorithms, Schulmann et al. \cite{equivalence_pg_soft_q} applies entropy regularization to Q-learning. They make an important observation about the equivalence between policy gradient and entropy regularized Q-learning.

To the best of our knowledge, no prior work has dealt with approximating the policy entropy for MDP with large multi-dimensional discrete action space. On the other hand, there have been many attempts to devise methods to encourage beneficial exploration for policy gradient. Nachum et al. \cite{urex} modifies the entropy term by adding weights to the log action probabilities, leading to a new optimization objective termed under-appreciated reward exploration. 




Dulac-Arnold et al. \cite{drl_large_discrete_action} embeds discrete actions in a continuous space, picks actions in the continuous space and map these actions back into the discrete space. However, their algorithm introduces a new hyper-parameter that requires tuning for every new task. Our approach involves no new hyper-parameter other than those normally used in deep learning. 

The LSTM policy parameterization can be seen as the adaptation of sequence modeling techniques in supervised machine learning, such as in speech generation \cite{wave_net} or machine translation \cite{lstm_translation} to reinforcement learning, as was previously done in \cite{actor_critic_sequence_prediction}.

\section{Conclusion}

In this paper, we developed several novel unbiased estimators for entropy bonus and its gradient. 
We did experimental work for two environments with large multi-dimensional action spaces. 
We found that the smoothed estimate of the entropy and the unbiased estimate of the entropy gradient can significantly increase performance with marginal additional computational cost. 

\section*{Appendix A. Hyperparameters}
\textit{Hyperparameters for hunter-rabbit game}

For IS, the numbers of hidden layers for smoothed entropy, unbiased gradient estimate, crude entropy and without entropy are 1, 1, 5 and 7 respectively. The entropy weights for smoothed entropy, unbiased gradient estimate and crude entropy are 0.03, 0.02 and 0.01 respectively. The hyper-parameters for smoothed mode entropy is not listed since the smoothed mode entropy equals the smoothed entropy for IS.

For CommNet, the numbers of communication step for without entropy, crude entropy, smoothed entropy and unbiased entropy gradient are 2, 2, 2 and 2 respectively. The sizes of the policy hidden layer for without entropy, crude entropy, smoothed entropy and unbiased entropy gradient are 256, 256, 256 and 128 respectively. The entropy weights for crude entropy, smoothed entropy and unbiased entropy gradient are 0.04, 0.04 and 0.01 respectively. The policies were optimized using Adam\cite{adam} with learning rate 3e-4. We found 3e-4 gives better performance than the learning rate 3e-3 originally used in \cite{comm_net}.

The LSTM policy has 128 hidden nodes.  For the MMDP policy, the number of hidden layers for smoothed entropy, smoothed mode entropy, unbiased gradient estimate, crude entropy and without entropy are 5, 3, 3, 4 3 and 3 respectively. Each MMDP layer has 128 nodes. 
We parameterize the baseline in \eqref{eq:pg_update}
with a FFN with one hidden layer of size 64. This network was trained using first visit Monte Carlo return to minimize the L1 loss between actual and predicted values of states visited during the episode. 

Both the policies and baseline are optimized after each episode with RMSprop \cite{rmsprops}. The RHS of \eqref{eq:pg_update} is clipped to $[-1, 1]$ before updating the policy parameters. The learning rates for the baseline, IS, LSTM and MMDP are $10^{-3}$, $10^{-3}$, $10^{-3}$, $10^{-4}$ respectively. 

To obtain the results in Table 1, the entropy weights for LSTM smoothed entropy, LSTM smoothed mode entropy, LSTM unbiased gradient estimate, LSTM crude entropy, MMDP smoothed entropy, MMDP smoothed mode entropy, MMDP unbiased gradient estimate and MMDP crude entropy are 0.02, 0.021, 0.031, 0.04, 0.02, 0.03,  0.03 and 0.01 respectively. 

To obtain the results in Table 2, the entropy weights for LSTM smoothed entropy, LSTM exact entropy, MMDP unbiased gradient estimate and MMDP exact entropy are 0.03, 0.01, 0.03 and 0.01 respectively. The MMDP networks have three layers with 128 nodes in each layer. Experimental results are averaged over five seeds (0-4).

\textit{Hyperparamters for Multi-Agent Multi-Arm Bandits}

The experiments were run with 4 agents and 10 arms. For the 10 arms, their rewards are $i$ for $i=1,\dots,10$. The LSTM policy has 32 hidden nodes. The baseline in (1) is a truncated average of the reward of the last 100 rounds. The entropy weight for crude entropy, smoothed entropy and unbiased gradient estimate are 0.005, 0.001 and 0.003 respectively. The learning rates for without entropy, crude entropy, smoothed entropy and unbiased gradient estimate are 0.006, 0.008, 0.002 and 0.005 respectively. Experimental results are averaged over ten seeds.

\section*{Appendix B. Proof of Theorem 2}
\textbf{Theorem 2. }\emph{	If $\prob(\act)$ has a multivariate normal distribution with mean and variance depending on $\theta$, then: 
	$$\entropym(\act) = \entropy \quad \forall \act \in \cA$$
	\\
Thus, the smoothed estimate of the entropy equals the exact entropy for a multivariate normal parameterization of the policy. }
\begin{proof}
We first note that for $\begin{bmatrix}\mathbf{X}_1 \\ \mathbf{X}_2\end{bmatrix}\sim N\left(\begin{bmatrix}\bmu_1 \\ \bmu_2\end{bmatrix},\begin{bmatrix}\bsigma_{11} & \bsigma_{12} \\ \bsigma_{21} & \bsigma_{22}\end{bmatrix}\right)$ where $\mathbf{X}_1$ and $\mathbf{X}_2$ are random vectors, we have $\mathbf{X}_2 \mid \mathbf{X}_1=\mathbf{x}_1\sim N(\bar{\bmu},\bar{\bsigma})$ where
\begin{align*}
\bar{\bmu} &= \bmu_2 + \bsigma_{21}\bsigma_{11}^{-1}(\mathbf{x}_1-\bmu_1) \\
\bar{\bsigma} &= \bsigma_{22} - \bsigma_{21}\bsigma_{11}\bsigma_{12}
\end{align*}
Observe that the covariance matrix of the conditional distribution does not depend on the value of $x_1$ \cite{applied_multi_stats}.

Also note that for $\mathbf{X}\sim N(\bmu,\bsigma)$, the entropy of $\mathbf{X}$ takes the form
\[
H(\mathbf{X})=\frac{k}{2}(\log 2\pi+1)+\frac{1}{2}|\bsigma|
\]
where $k$ is the dimension of $\mathbf{X}$ and $|\cdot|$ denotes the determinant. Therefore, the entropy of a multivariate normal random variable depends only on the variance and not on the mean. 

Because $\A$ is multivariate normal, the distribution of $A_i$ given $A_1 = a_1,\dots,A_{i-1} = a_{i-1}$ has a normal distribution with a variance $\sigma_i^2$ that does not depend on $a_1,\dots,a_{i-1}$. Therefore 
\[
H_\theta(A_i | a_1,\ldots,a_{i-1}) = \frac{1}{2}(\log 2 \pi + 1 +\sigma_i^2)
\]
does not depend on $a_1,\dots,a_{i-1}$ and hence $\entropym(\act)$ does not depend on $\act$.  Combining this with the fact that $\entropym(\act)$ is an unbiased estimator for $\entropy$ gives
$\entropym(\act) = \entropy$ for all $\act \in \cA$. 
\end{proof}

\section*{Appendix C. Proof of Theorem 4}
\textbf{Theorem 4. }\emph{ If $\act$ is a sample from $\prob(\cdot)$, then
$$ \derp \entropym(\act) + \sum_{i=1}^d H_\theta^{(i)} (\act_{i-1}) 
\derp \sum_{j=1}^{i-1} \lp (a_j| \act_{j-1})$$
is an unbiased estimator for the gradient of the entropy.}
\begin{proof}
From Equation\eqref{der_entropy_expansion}, we have:
\begin{equation}
	\derp \entropy  = - \sumtod{i} \sumtod{j} \EofA [ \lpA \derp \lpAj ]
    \label{entropy_gradient_sum_exp}
\end{equation}
We will now use conditional expectation to calculate the terms in the double sum.

For $i < j$:
\begin{align*}
	& \EofA[\lpA \derp \lpAj | \A_{j-1}] \\
	= & \lpA \EofA [\derp \lpAj | \A_{j-1}] = 0
\end{align*}
For $i > j$:
\begin{align*}
	&\EofA [\lpA \derp \lpAj | \A_{i-1}] \\
	= & \derp \lpAj \EofA [\lpA | \A_{i-1} ] \\
	= & - \derp \lpAj H_\theta^{(i)} (\A_{i-1})
\end{align*}
For $i = j$:
\begin{align*}
\EofA [\lpA \derp \lpA | \A_{i-1}] = - \derp H_\theta^{(i)} (\A_{i-1})
\end{align*}

Combining these three conditional expectations with (\ref{entropy_gradient_sum_exp}), we obtain:
\begin{align*}
	\derp \entropy = \EofA[ \derp \entropym(\A)  + \sum_{i=1}^d H_\theta^{(i)} (\A_{i-1}) \derp \sum_{j=1}^{i-1} \lp (A_j| \A_{j-1})]
\end{align*}

Alternatively, Theorem 4 could also be proven by applying Theorem 1 of \cite{schulman2015gradient} . 
\end{proof}

 \section*{Appendix D. State Representation For CommNet}
Sukhbaatar et al.\cite{comm_net} proposes CommNet to handle multi-agent environments where each agent observes only part of the state and the number of agents changes throughout an episode. We thus modify the state representation of the hunters and rabbits environment to better reflect the strengths of CommNet. Each hunter only sees its own id, its position and the positions of all rabbits. More precisely, the state each hunter receives is [{\it hunter id}, {\it hunter position}, {\it all rabbit positions}]. 

\subsubsection*{Acknowledgements}
We would like to thank Martin Arjovsky for his input and suggestions at both the early and latter stages of this research. Our gratitude also goes to the HPC team at NYU, NYU Shanghai, and NYU Abu Dhabi.

\bibliographystyle{unsrt}
\bibliography{entropy_nips}

\end{document}